# HCOMC: A Hierarchical Cooperative On-Ramp Merging Control Framework in Mixed Traffic Environment on Two-Lane Highways

Tianyi Wang[1†], Yangyang Wang[2], Jie Pan[3], Junfeng Jiao[4], Christian Claudel[1]

*Abstract*—Highway on-ramp merging areas are common bottlenecks to traffic congestion and accidents. Currently, a cooperative control strategy based on connected and automated vehicles (CAVs) is a fundamental solution to this problem. While CAVs are not fully widespread, it is necessary to propose a hierarchical cooperative on-ramp merging control (HCOMC) framework for heterogeneous traffic flow on two-lane highways to address this gap. This paper extends longitudinal car-following models based on the intelligent driver model and lateral lane-changing models using the quintic polynomial curve to account for human-driven vehicles (HDVs) and CAVs, comprehensively considering human factors and cooperative adaptive cruise control. Besides, this paper proposes a HCOMC framework, consisting of a hierarchical cooperative planning model based on the modified virtual vehicle model, a discretionary lane-changing model based on game theory, and a multi-objective optimization model using the elitist non-dominated sorting genetic algorithm to ensure the safe, smooth, and efficient merging process. Then, the performance of our HCOMC is analyzed under different traffic densities and CAV penetration rates through simulation. The findings underscore our HCOMC's pronounced comprehensive advantages in enhancing the safety of group vehicles, stabilizing and expediting merging process, optimizing traffic efficiency, and economizing fuel consumption compared with benchmarks.

## I. INTRODUCTION

Highway on-ramp merging areas are common bottlenecks to traffic congestion and accidents. Notably, the speed differences between mainline vehicles and on-ramp vehicles often lead to traffic delays. With the development of automotive technology, cooperative control based on connected and automated vehicles (CAVs) has emerged as a fundamental solution to improve vehicle safety and alleviate traffic congestion [1]. However, due to current technological and economic constraints, CAVs cannot completely replace traditional human-driven vehicles (HDVs) in the near term [2]. Consequently, how to control CAVs' motion states in the cooperative planning area considering the mixed traffic flow involving HDVs and CAVs, thereby ensuring merging safety and traffic efficiency, becomes a key challenge to tackle.

The feedback control method based on the virtual vehicle theory was initially introduced by Uno et al. [3], involving mapping the on-ramp merging vehicles to the main lane based on their distance from a fixed merging point and implementing speed control. Building upon this, Milanés et al.[4], [5] proposed a safe distance control method to optimize the longitudinal motion of mainline vehicles to reserve a desired gap for on-ramp vehicles to merge. Wang et al.[6], [7] mapped on-ramp vehicles to the same main lane to ensure safe car-following behaviors and constructed a feedback function including distance error and speed error to achieve cooperative control within a given on-ramp merging control region. Despite these efforts, virtual vehicle theory may not align with actual traffic dynamics without considering the speed differences between the vehicles in different lanes. Besides, while most research only analyzed traffic flow performance on single main-lane highways, neglecting the potential for lateral cooperation among vehicles on multi-lane highways and simplifying lateral behavior details make it unconvincing to derive the comprehensive advantages of CAV technologies and inevitably wastes traffic capacity.

In addition to the virtual vehicle theory, previous studies investigated CAV-based cooperative control incorporating game theory [8], optimal control [9], and reinforcement learning[10]. Ntousakis et al. [11] established two merging sequence decision rules: first in first out (FIFO) and linear prediction of vehicle speed, which can ensure safety, but had limited adaptability to real-world traffic characteristics. Building on the FIFO model, Jing et al. [12] introduced a game mechanism to establish a decision model for merging sequences, which improved fuel economy, passenger comfort, and overall traffic efficiency by using the Pareto optimal algorithm. Moreover, Wang et al. [13] proposed an optimal cooperative control method for on-ramp merging on multi-lane highways, improving traffic stability, speed, and efficiency of traffic flow. Yu et al. [14] integrated interactive Monte Carlo tree search with deep reinforcement learning, aiming to enhance interaction rationality, efficiency and safety of CAVs. Most studies were based on pure CAV traffic scenarios without the consideration of actual complex mixed traffic environments involving HDVs. Meanwhile, only a few studies considered overall traffic attributes, i.e. safety, stability, economy, comfort and efficiency.

Recently, several methods have been proposed for highway on-ramp merging problems in mixed traffic environment. Rios-Torres et al. [15] proposed an on-ramp merging control architecture in mixed traffic environments, where HDVs were built based on the Gips model and CAVs' trajectories were planned by an unconstrained optimal control method, which guided CAVs to arrive at a specified merging point according to a pre-defined merging time. Sun et al. [16] utilized the Gipps model and the intelligent driver model (IDM) to characterize HDVs and CAVs, respectively, and then they used a rule-based approach for trajectory planning. The above research ignored the uncertainty of HDVs' behaviors, which was not in line with the actual characteristics of human drivers. To address this, Liao et al. [17], [18] designed cooperative rules based on game theory, and verified the effectiveness of the proposed merging method through software simulation and digital twin experiments. Hou et al. [19] proposed a CORMC model to ensure efficient and safe merging of the vehicles in the multi-lane merging zone under mixed traffic environments. However, most current studies assumed that HDVs' behaviors strictly conformed to a specific vehicle model, and the degree of HDV cooperation was expected to be too high.

†Corresponding author: Tianyi Wang.

[1]Tianyi Wang and Christian Claudel are with the Department of Civil, Architectural, and Environmental Engineering, The University of Texas at Austin, Austin, TX 78712, USA. Email: bonny.wang@utexas.edu; christian.claudel@utexas.edu.

[2]Yangyang Wang is with the School of Automotive Studies, Tongji University, Shanghai 201804, China. Email: wyangyang@tongji.edu.cn.

[3]Jie Pan is with the Department of Civil Engineering, Tsinghua University, Beijing 100084, China. Email: panj21@mails.tsinghua.edu.cn.

[4]Junfeng Jiao is with the School of Architecture, The University of Texas at Austin, Austin, TX 78712, USA. Email: jjiao@austin.utexas.edu.

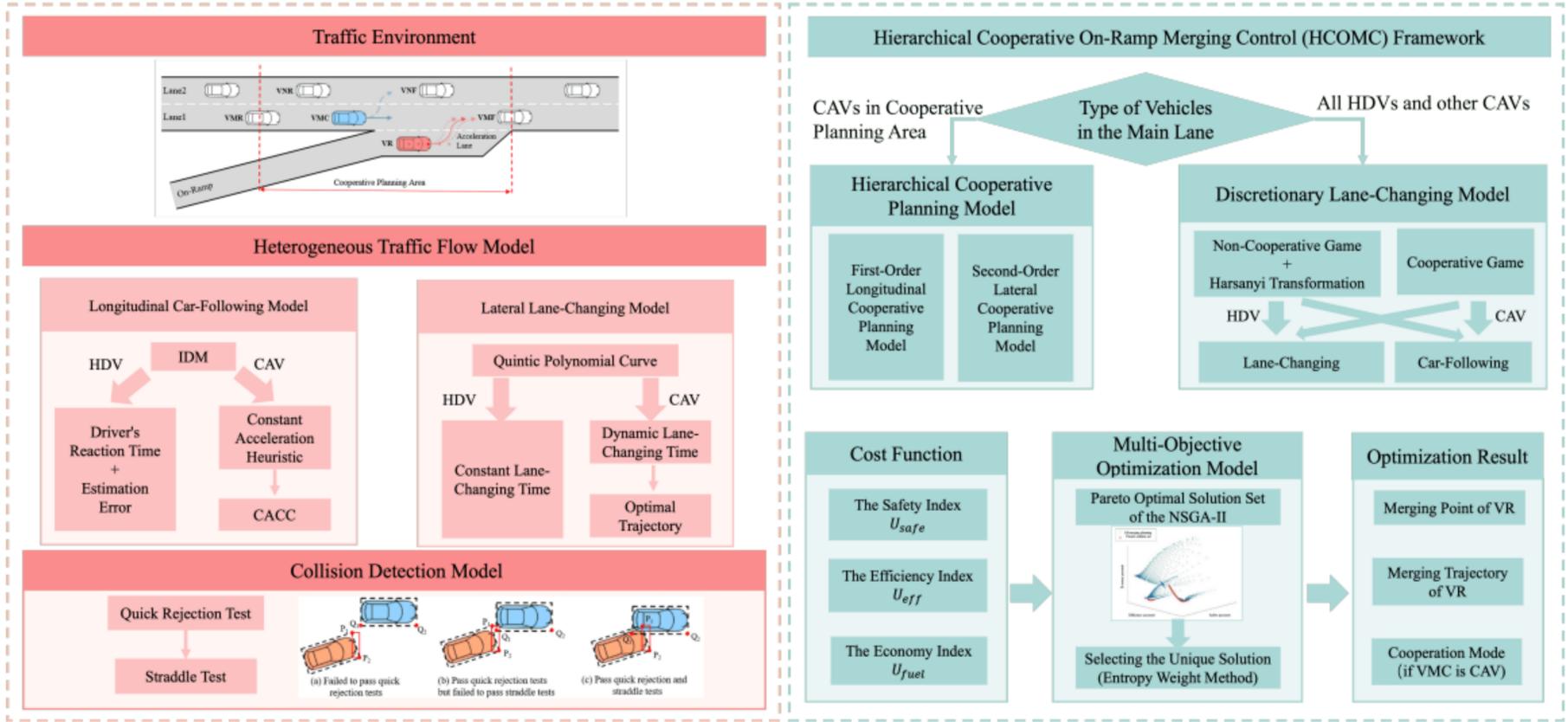

Fig. 1. Framework of the proposed hierarchical cooperative on-ramp merging control (HCOMC) framework

According to the analysis above, this paper makes three contributions to existing research:

- This paper modifies the longitudinal car-following models and lateral lane-changing models to capture the distinct driving characteristics of mixed traffic flow on two-lane highways.
- This paper proposes HCOMC (Figure 1), a framework making up of the hierarchical cooperative planning model, the discretionary lane-changing model, and the multi-objective optimization model to ensure the safe, smooth, and efficient merging process.
- The simulation is conducted to verify the effectiveness of our HCOMC under different typical working conditions, considering varying traffic densities and CAV penetration rates, and comprehensive evaluation indexes are designed to prove our model's overall optimality.

## II. PROBLEM FORMULATION

### A. Traffic Environment

This paper focuses on a two-lane highway on-ramp merging area, which consists of six key vehicles as shown in Figure 1, i.e. vehicle on ramp (VR), vehicle in lane 1 to cooperate VR (VMC), vehicle in lane 1 in front of VMC (VMF), vehicle in lane 1 in rear of VMC (VMR), vehicle in lane 2 in front of VMC (VNF), and vehicle in lane 2 in rear of VMC (VNR).

### B. Heterogeneous Traffic Flow Model

*1) Longitudinal Car-Following Model:* IDM is adopted as the basic model to construct the car-following models of HDVs and CAVs.

$$\begin{cases} a_n(s_n, v_n, \Delta v_n) = a\left(1 - \left(\frac{v_n}{v_0}\right)^\delta - \left(\frac{s^*(v_n, \Delta v_n)}{s_n}\right)^2\right) \\ s^*(v_n, \Delta v_n) = s_0 + \max\left(0, v_n T_s + \frac{v_n \cdot \Delta v_n}{2\sqrt{ab}}\right) \end{cases} \quad (1)$$

Where, $a_n$ is the desired acceleration; $s_n$ is the actual car-following distance; $v_n$ is the actual speed; $\Delta v_n$ means the actual speed difference between the subject vehicle and its leading vehicle; $a$ means the maximum desired acceleration; $v_0$ is the desired speed; $\delta$ means the acceleration index; $s_0$ represents the desired car-following distance; $s^*(v_n, \Delta v_n)$ is the static safety distance; $T_s$ is the safety time headway; and $b$ is the comfort deceleration.

a) Human-Driven Vehicle: This paper introduces human driver's reaction time and estimation error to IDM. According to our previous work [20], the HDV longitudinal car-following model incorporating reaction delay time and estimation error is expressed as:

$$a_n^{\text{HDV}} = f\left(s_n^{\text{est}}(t - \tau_{s_n}), v_n(t - \tau_{v_n}), \Delta v_n^{\text{est}}(t - \tau_{\Delta v_n})\right) \quad (2)$$

Where, $s_n^{est}$ is the estimated car-following distance; $\Delta v_n^{est}$ is the estimated speed difference; and $\tau_{s_n}$, $\tau_{v_n}$ and $\tau_{\Delta v_n}$ are the reaction time with respect to the actual car-following distance, the actual speed and the actual speed difference.

b) Connected and Automated Vehicle: This paper integrates the constant acceleration heuristic into IDM to account for CAVs. According to our previous work [20], the CAV longitudinal car-following model is expressed as:

$$a_n^{CAV} = (1-c)a_{IDM} + c\left(a_{CAH} + b\tanh\left(\frac{a_{IDM} - a_{CAH}}{b}\right)\right) \quad (3)$$

Where, $a_{CAH}$ denotes the maximum acceleration leading to no crashes; $a_{IDM}$ is the acceleration of the traditional IDM; and $c$ is the cooling factor.

*2) Lateral Lane-Changing Model:* Quintic polynomial curve is adopted as the basic model to construct the lane-changing trajectories of HDVs and CAVs.

$$y(x) = a_0 + a_1 x + a_2 x^2 + a_3 x^3 + a_4 x^4 + a_5 x^5, \quad x \in [x_0, x_f] \quad (4)$$

Where, $y$ is the lateral displacement; $a_i$ is the polynomial corresponding coefficient; $x$ is the longitudinal displacement; and $x_0$ and $x_f$ are longitudinal displacement at the beginning and end of lane change, respectively.

a) Human-Driven Vehicle: This paper introduces the HDV longitudinal car-following model in Equation (2) to the lateral lane-changing model:

$$\begin{cases} x^{HDV} = x_0 + u(t-t_0) + \int_{t_0}^{t_f}(\int_{t_0}^{t_f} a_n^{HDV} dt) dt \\ y^{HDV} = y(x^{HDV}) \end{cases} \quad (5)$$

Where, $x^{HDV}$ and $y^{HDV}$ represent the longitudinal and lateral displacement of HDV, respectively; $t_0$ and $t_f$ are the beginning and end moment of lane change, respectively; and $u$ is the longitudinal speed at the beginning of lane change.

Referring to the lane-changing time on the common high-speed condition [21], this paper sets the HDV lane-changing time $(t_f - t_0)$ as a constant, which is 4 seconds.

b) Connected and Automated Vehicle: This paper introduces the CAV longitudinal car-following model in Equation (3) to the lateral lane-changing model:

$$\begin{cases} x^{CAV} = x_0 + u(t-t_0) + \int_{t_0}^{t_f}(\int_{t_0}^{t_f} a_n^{CAV} dt) dt \\ y^{CAV} = y(x^{CAV}) \end{cases} \quad (6)$$

Where, $x^{CAV}$ and $y^{CAV}$ represent the longitudinal and lateral displacement of CAV, respectively.

The CAV lane-changing time is dynamic, which can be obtained by the HCOMC framework introduced later.

### C. Collision Detection Model

The collision scenarios include rear-end collision and side-impact collision. To streamline the computation process, this paper leverages computer graphics methods and implements a rapid collision judgment scheme based on the principles of quick rejection tests and straddle tests.

*1) Quick Rejection Test:* Assume that $P_1P_2$ and $Q_1Q_2$ are the two potential collision boundaries within the rectangular vehicle model. The purpose of the quick rejection test is to provide a preliminary assessment to exclude the case that $P_1P_2$ and $Q_1Q_2$ do not intersect, which would indicate that the two vehicles are not involved in a collision situation. If the quick rejection test is passed, it suggests that $P_1P_2$ and $Q_1Q_2$ may intersect, thereby necessitating the subsequent execution of the straddle tests to determine whether the two-line segments intersect, which may lead to collisions.

*2) Straddle Test:* When $P_1P_2$ intersects with $Q_1Q_2$, it means that the endpoints $P_1$ and $P_2$ lie on opposite sides of the line where $Q_1Q_2$ is located, while $Q_1$ and $Q_2$ lie on opposite sides of the line where $P_1P_2$ is located. This judgment can be achieved by calculating the cross product:

$$\begin{cases} (P_1Q_1 \times Q_1Q_2) \cdot (P_2Q_1 \times Q_1Q_2) < 0 \\ (Q_1P_1 \times P_1P_2) \cdot (Q_2P_1 \times P_1P_2) < 0 \end{cases} \quad (7)$$

By identifying the dangerous boundaries in different collision scenarios and applying the quick rejection tests and straddle tests, collision safety can be determined. And for each feasible merging trajectory of VR, the merging sequence is then determined based on the collision safety between VR and its surrounding vehicles, i.e. VMC and VMR in Figure 1.

## III. METHODOLOGY

### A. Hierarchical Cooperative Planning Model

When VMC is a CAV, a hierarchical cooperative planning model is established for the two-lane highway on-ramp merging area, including a first-order longitudinal and a second-order lateral cooperative planning model.

*1) First-Order Longitudinal Cooperative Planning Model:* In this paper, an improved virtual vehicle model in main lane 1 based on traffic flow motion boundaries is proposed.

$$\begin{cases} x_{VV1}(t_0') = x_{VMF}(t_0') \\ v_{VV1}(t_0') = v_{VMF}(t_0') \\ x_{VV1}(t_f') = x_{VR}(t_f') \\ v_{VV1}(t_f') = v_{VR}(t_f') \end{cases} \quad (8)$$

Where, $x_{VV1}$ and $v_{VV1}$ represent the displacement and speed of the virtual vehicle of VR; and $t_0'$ and $t_f'$ are the beginning and end moment of VR planning.

The longitudinal motion trajectory equation and the speed equation are shown as follows:

$$\begin{pmatrix} x_{VMF}(t_0') & x_{VR}(t_f') \\ v_{VMF}(t_0') & v_{VR}(t_f') \end{pmatrix} = \begin{pmatrix} b_{11} & b_{12} & b_{13} & b_{14} \\ b_{12} & 2b_{13} & 3b_{14} & 0 \end{pmatrix} \begin{pmatrix} 1 & 1 \\ t_0' & t_f' \\ t_0'^2 & t_f'^2 \\ t_0'^3 & t_f'^3 \end{pmatrix} \quad (9)$$

The undetermined coefficients $b_{ij}$ in the state equation can

be obtained by solving:

$$\begin{pmatrix} b_{11} \\ b_{12} \\ b_{13} \\ b_{14} \end{pmatrix} = \begin{pmatrix} 1 & t_0' & t_0'^2 & t_0'^3 \\ 1 & t_f' & t_f'^2 & t_f'^3 \\ 0 & 1 & 2t_0' & 3t_0'^2 \\ 0 & 1 & 2t_f' & 3t_f'^2 \end{pmatrix}^{-1} \begin{pmatrix} x_{VMF}(t_0') \\ x_{VR}(t_f') \\ v_{VMF}(t_0') \\ v_{VR}(t_f') \end{pmatrix} \quad (10)$$

After the longitudinal motion trajectory equation and the speed equation are obtained during the cooperative planning process of the virtual vehicle of VR, the acceleration variation equation of the virtual vehicle of VR can be calculated by derivation. Then, according to the heterogeneous traffic flow model, the longitudinal acceleration of the virtual vehicle of VR can be obtained.

*2) Second-Order Lateral Cooperative Planning Model:* Similar to the longitudinal cooperative planning model, the motion planning equations of the virtual vehicle of VMC in main lane 2 can be obtained:

$$\begin{cases} x_{VV2}(t_0'') = x_{VNF}(t_0'') \\ v_{VV2}(t_0'') = v_{VNF}(t_0'') \\ x_{VV2}(t_f'') = x_{VMC}(t_f'') \\ v_{VV2}(t_f'') = v_{VMC}(t_f'') \end{cases} \quad (11)$$

$$\begin{pmatrix} x_{VNF}(t_0'') & x_{VMC}(t_f'') \\ v_{VNF}(t_0'') & v_{VMC}(t_f'') \end{pmatrix} = \begin{pmatrix} c_{11} & c_{12} & c_{13} & c_{14} \\ c_{12} & 2c_{13} & 3c_{14} & 0 \end{pmatrix} \begin{pmatrix} 1 & 1 \\ t_0'' & t_f'' \\ t_0''^2 & t_f''^2 \\ t_0''^3 & t_f''^3 \end{pmatrix} \quad (12)$$

$$\begin{pmatrix} c_{11} \\ c_{12} \\ c_{13} \\ c_{14} \end{pmatrix} = \begin{pmatrix} 1 & t_0'' & t_0''^2 & t_0''^3 \\ 1 & t_f'' & t_f''^2 & t_f''^3 \\ 0 & 1 & 2t_0'' & 3t_0''^2 \\ 0 & 1 & 2t_f'' & 3t_f''^2 \end{pmatrix}^{-1} \begin{pmatrix} x_{VNF}(t_0'') \\ x_{VMC}(t_f'') \\ v_{VNF}(t_0'') \\ v_{VMC}(t_f'') \end{pmatrix} \quad (13)$$

Where, $x_{VV2}$ and $v_{VV2}$ represent the displacement and speed of the virtual vehicle of VMC; and $t_0''$ and $t_f''$ are the beginning and end moment of VMC planning.

During the merging process, VMC changes lane and its leading vehicle shifts from VMF to VNF. The state differences between VMF and VNF trigger a sudden jump in the output of the IDM-based model. Therefore, a hyperbolic tangent transition function is introduced to achieve a smooth lane-changing maneuver:

$$\begin{cases} a_{tar} = \Psi(\tau)a_{new} + (1 - \Psi(\tau))a_{ori} \\ \Psi(\tau) = \frac{1}{2}[\tanh(\lambda\tau - \gamma) + 1] \end{cases} \quad (14)$$

Where, $a_{tar}$ represents the desired acceleration with transition function; $a_{ori}$ and $a_{new}$ are the desired acceleration before and after lane change, respectively; $\Psi(\tau)$ is the transition function, which satisfies $\Psi(t_0) = 0$ and $\Psi(t_f) = 1$; $\tau$ means the timing of lane change; and $\gamma$ and $\lambda$ are the parameters reflecting the phase and speed of the transition, respectively.

### B. Discretionary Lane-Changing Decision Model

Discretionary lane-changing behavior is one of the most common highway operations, which seriously affects traffic efficiency and safety. Generally, the subject vehicle (SV) in the main lane has priority in making driving decisions as it encounters traffic situations earlier than its follower vehicle (FV) in the adjacent lane, and its decisions are rarely restrained by FV. Therefore, the interactions between SV and FV can be modeled as a two-player Stackelberg game with SV as the leader. For SVs with lane-changing intentions, their action set includes changing lanes and keeping car-following. FVs, according to varying driving styles, have actions including changing lane and keeping car-following (i.e., at constant speed, accelerating and decelerating).

Initially, assuming SV as a CAV, it gathers essential information from its surrounding environment using cooperative adaptive cruise control (CACC) technologies. Subsequently, SV and FV engage in a strategic game. Applying Stackelberg game principles and Harsanyi transformation theory, the payoff functions are computed across various action combinations, aiming to identify an optimal action for both SV and FV. In scenarios where SV conducts lane changes, dynamic safety domains and optimal lane-changing trajectories are calculated to enhance collision safety and lateral stability [20]. Simultaneously, if FV is also a CAV, modified virtual vehicle models are adopted, otherwise transition models are accepted, in order to improve cooperation between CAVs, thereby enhancing traffic efficiency and ride comfort. According to the idea of mixed-strategy Nash equilibrium, SV takes the minimization of the expectation of its own payoff functions as the optimal strategy, and the combination of the strategy space and the mixed-strategy probability distribution of FV can be used to compute the expected payoff functions under different strategies (i.e. change lane and not changing lane) of SV.

### C. Multi-Objective Optimization Model

In the cooperative planning model for highway on-ramp merging areas presented in this paper, three primary aspects require optimization: (1) the optimal merging position of VR; (2) the merging trajectory of VR; (3) the cooperation mode of VMC. Consequently, the optimal cooperative planning approach proposed in this paper can be framed as a multi-objective optimization problem.

*1) Optimization Objectives:* According to the optimization problem, this paper sets three key evaluation indexes to form cost functions: safety, economy, and efficiency.

*a) **Safety***: The critical acceleration, which is used as the safety evaluation criterion at the end of on-ramp merging, represents the minimum acceleration required for the rear vehicle to avoid rear-end collision when the front vehicle suddenly decelerates in an emergency. Thus, the safety index $U_{safe}$ is set to the critical acceleration, satisfying:

$$U_{safe} = a_{cr} = \frac{v_r^2}{2(d_{cri} - D - v_r T + \frac{v_f^2}{2a_{merg}})} \quad (15)$$

Where, $a_{cr}$ is the critical acceleration; $v_r$ and $v_f$ are the speed of the subject vehicle and its leading vehicle, respectively; $T$ is the delay time; $a_{merg}$ is the acceleration in emergency; $D$ is the minimum distance; and $d_{cri}$ is the critical distance.

b) **Economy**: The economy index $U_{fuel}$ is defined as the total fuel consumption of VMC and VR throughout the merging process. The fuel consumption rate is obtained:

$$\begin{cases} \dot{f}e = \dot{f}_{cruise} + \dot{f}_{accel} \\ \dot{f}_{cruise}(t) = [1 \; v(t) \; v(t)^2 \; v(t)^3] Q^T \\ \dot{f}_{accel}(t) = a(t)[1 \; v(t) \; v(t)^2] R^T \\ U_{fuel} = \int_{t_0}^{t_f} \dot{f}e_{VR}(t)dt + \int_{t_0}^{t_f} \dot{f}e_{VMC}(t)dt \end{cases} \quad (16)$$

Where, $\dot{f}e$ denotes the fuel consumption rate; $Q$ and $R$ are the fuel consumption parameter matrix for cruising and acceleration, respectively; and $\dot{f}_{cruise}$ and $\dot{f}_{accel}$ represent the fuel consumption rates during cruising and acceleration.

c) **Efficiency**: For the efficiency index $U_{eff}$ in merging process, we refer to the acceleration incentive model:

$$U_{eff} = a_{(VR)new} - a_{(VR)ori} + \eta \cdot \sum_{i=1}^{N}(a_{(i)new} - a_{(i)ori}) \quad (17)$$

Where $a_{(i)new}$ and $a_{(i)ori}$ are the acceleration before and after lane change, respectively, and $i$ represents different vehicles, i.e. VMC, VNR and VMR; and $\eta$ is the politeness factor, determining to which degree these vehicles influence the lane-changing decision.

2) *Optimization Algorithm*: Elitist non-dominated sorting genetic algorithm (NSGA-II) can quickly find the Pareto boundary while maintaining the diversity of the population. In this paper, NSGA-II is adopted due to its high efficiency, good real-time performance, and strong versatility in optimization algorithm. The expression of the multi-objective optimization model is shown in:

$$\begin{cases} \min F(x) = (f_1(x), f_2(x), ..., f_m(x)) \\ s.t. \; x \in \Omega \end{cases} \quad (18)$$

Where, $\Omega$ is the decision space; and $F(x)$ is the target space for the spatial domain conversion from $\Omega$ to $R^m$.

In the multi-objective optimization problem described above, if there is a solution $x^* \in \Omega$, and meanwhile, there is no other solution to make $F(x)$ dominate $F(x^*)$, then $x^*$ can be called the Pareto optimal solution of the equation, and $F(x^*)$ can be called the Pareto optimal vector.

After getting the Pareto optimal solution set of NSGA-II, the unique optimal solution needs to be picked out as the final output of the optimization model. The steps to select the unique solution from the Pareto optimal solution set are as follows: (1) If $U_{safe} > 4$, then, select the merging planning with the smallest $U_{safe}$. (2) If $U_{safe} \leq 4$, normalize the efficiency cost and the economy cost in the merging solutions of $U_{safe} \leq 4$, and sum up the normalized results to select the unique optimal solution.

$$U = \frac{U_{eff} - \min(U_{eff})}{\max(U_{eff}) - \min(U_{eff})} + \frac{U_{fuel} - \min(U_{fuel})}{\max(U_{fuel}) - \min(U_{fuel})} \quad (19)$$

## IV. EXPERIMENTS

The superiority of the NSGA-II model is verified by comparing it with other benchmarks, i.e. the particle swarm optimization (PSO) model [22] and the simulated annealing (SA) model [23]. Furthermore, simulations conducted across multiple traffic densities and CAV penetration rates validate the comprehensive advantages of the proposed HCOMC framework compared to the FIFO model [11] and the game theory model [17]. Specific typical working conditions are displayed in TABLE I. The simulation results of different multi-objective optimization models and on-ramp merging control models are shown in TABLE II and TABLE III.

TABLE I
PARAMETERS IN SPECIFIC SIMULATION CONDITIONS

| Number | Average Time Headway in Main Lane 1 (s) | Average Time Headway in Main Lane 2 (s) | CAV Penetration Rate (/) |
|---|---|---|---|
| Condition 1 | 5 | 5 | 90% |
| Condition 2 | 5 | 7 | 90% |
| Condition 3 | 5 | 3 | 90% |
| Condition 4 | 5 | 5 | 60% |
| Condition 5 | 5 | 5 | 30% |

TABLE II
RESULTS OF DIFFERENT MULTI-OBJECTIVE OPTIMIZATION MODELS

| Number | Model | Crit. Dist. (m) | Aver. Acc. (m/s²) | Stab. Time (s) | LSRV (m²) | Fuel Cons. (L) |
|---|---|---|---|---|---|---|
| Condition 1 | PSO | 108.37 | 0.0244 | 19.3 | 3.9024 | 8.0433 |
| | SA | 108.45 | 0.0243 | 19.1 | 3.9021 | 8.0424 |
| | **NSGA-II** | **114.21** | **0.0225** | **16.6** | **3.8879** | **8.0375** |
| Condition 2 | PSO | 150.97 | 0.0475 | 9.3 | 3.0009 | 8.3909 |
| | SA | 150.97 | 0.0475 | 9.3 | 3.0009 | 8.3937 |
| | **NSGA-II** | **149.17** | **0.0477** | **11.8** | **2.9996** | **8.3982** |
| Condition 3 | PSO | 106.18 | 0.0517 | 19.5 | 5.7177 | 8.0780 |
| | SA | 106.19 | 0.0517 | 19.7 | 5.7177 | 8.0779 |
| | **NSGA-II** | **112.01** | **0.0494** | **17.2** | **5.7021** | **8.0442** |

### A. Safety of Group Vehicles

The most **critical distance** (Crit. Dist.) is chosen as the evaluation index of the safety of group vehicles. Under Condition 1 and Condition 3, where VMC implements longitudinal cooperation, NSGA-II can improve the traffic safety significantly, which increases by 5.39%, 5.31% and 5.49%, 5.48% compared with PSO and SA, respectively. When VMC implements lateral cooperation under Condition 2, although NSGA-II performs inferior to the other two models, the difference is only about 1%. It can be seen in TABLE III that our HCOMC improves the safety of group vehicles in all conditions. Compared to the FIFO model and the game theory model, the proposed HCOMC improves the critical distance by 9.11% and 5.13%, respectively. What is noteworthy is that under Condition 2, where VMC implements longitudinal cooperation both in FIFO and game theory, while implementing lateral cooperation in our HCOMC, the maximum increase rate can reach over 46%.

TABLE III
RESULTS OF DIFFERENT ON-RAMP MERGING CONTROL MODELS

| Number | Model | Crit. Dist. (m) | Aver. Acc. (m/s²) | Stab. Time (s) | LSRV (m²) | Fuel Cons. (L) |
|---|---|---|---|---|---|---|
| Condition 1 | FIFO | 104.67 | 0.0267 | 22.5 | 3.9122 | 8.0660 |
| | Game | 108.64 | 0.0242 | 19.0 | 3.9016 | 8.0409 |
| | **HCOMC** | **114.21** | **0.0225** | **16.6** | **3.8879** | **8.0375** |
| Condition 2 | FIFO | 85.47 | 0.0442 | 26.1 | 3.1131 | 8.1516 |
| | Game | 80.33 | 0.0513 | 25.4 | 3.1444 | 8.0598 |
| | **HCOMC** | **149.17** | **0.0477** | **11.8** | **2.9996** | **8.3982** |
| Condition 3 | FIFO | 102.30 | 0.0545 | 22.9 | 5.7286 | 8.1057 |
| | Game | 106.53 | 0.0515 | 19.2 | 5.7168 | 8.0744 |
| | **HCOMC** | **112.01** | **0.0494** | **17.2** | **5.7021** | **8.0442** |
| Condition 4 | FIFO | 104.67 | 0.0267 | 22.5 | 3.9125 | 8.0510 |
| | Game | 108.64 | 0.0242 | 19.0 | 3.9019 | 8.0220 |
| | **HCOMC** | **114.21** | **0.0225** | **16.6** | **3.8882** | **8.0148** |
| Condition 5 | FIFO | 125.62 | 0.0427 | 20.4 | 3.9506 | 8.0093 |
| | Game | 115.68 | 0.0482 | 23.1 | 3.9727 | 7.9763 |
| | **HCOMC** | **129.99** | **0.0360** | **18.8** | **3.9494** | **7.8927** |

### B. Stability and Rapidity of Merging

In this paper, the **average acceleration** (Aver. Acc.) of the six key vehicles and the **time required to stabilize after merging** (Stab. Time) are selected as the stability and rapidity evaluation indexes. Almost under all the working conditions, the proposed HCOMC improves both of the evaluation indexes, which indicates the great improvements in rapidity and stability of merging. Under Condition 2, compared with FIFO and game theory, the proposed HCOMC model can shorten the time to stabilize after merging by 54.79% and 53.53%, respectively, benefiting from the lateral cooperation mode of VMC at the expense of stability due to additional lane-changing behaviors. The results under Condition 1, Condition 4, and Condition 5 indicate that the performance of our HCOMC maintains superiority and stable under different CAV penetration rates.

### C. Efficiency of Traffic Flow

This paper uses **low-speed region volume** (LSRV) as the evaluation index of traffic efficiency. The low-speed region volume is computed by integrating the product of velocity, longitudinal displacement, and time. Under Condition 1 and Condition 3, where VMC implements longitudinal cooperation, NSGA-II reduces the low–speed region volume by

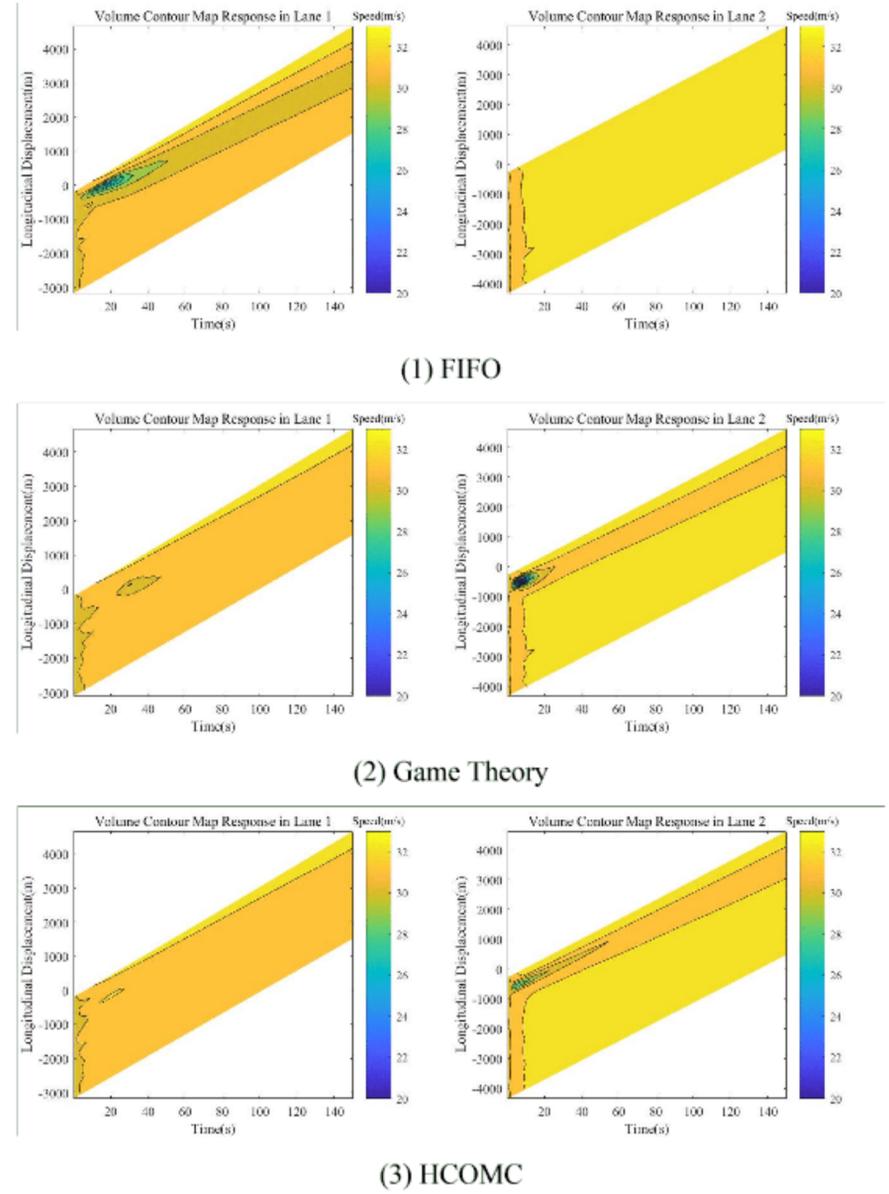

(1) FIFO

(2) Game Theory

(3) HCOMC

Fig. 2. Volume contour map response of two lanes under Condition 2

0.37%, 0.36% and both 0.27% compared with PSO and SA, respectively. In contrast, VMC implements lateral cooperation under Condition 2, where NSGA-II also performs the best in this case. The volume contour maps in Figure 2 shows that our HCOMC has smoother speed fluctuations and a smaller impact time frame in the mode of lateral cooperation (Condition 2), indicating that our HCOMC outperforms other models in traffic efficiency. Simulation results also showcase that the decreases in the number of CAVs lead to higher total low-speed region volumes, which indicates the significant role of CAVs in improving the efficiency of traffic flow.

### D. Fuel Consumption Economy

In this paper, the **total fuel consumption** (Fuel Cons.) of the traffic flow is utilized as the evaluation index for fuel consumption economy. Under Condition 1 and Condition 3, NSGA-II outperforms the other two optimization algorithms, which indicates the advantages of NSGA-II in improving the overall fuel consumption performance of traffic flow. Under Condition 2, NSGA-II performs inferior to the other two models, but the difference is only about 1%. Our HCOMC reduces the total fuel consumption by 0.35% and 0.04%, compared with the FIFO model and the game theory model under Condition 1. However, under Condition 2, the fuel consumption of the HCOMC is larger than that of the FIFO model and the game theory model, which is attributed to the additional lateral cooperation maneuvers of VMC.

*E. Discussion*

Results show that NSGA-II consistently performs best under almost all the working conditions, particularly in longitudinal cooperation mode. This demonstrates that the NSGA-II model is more effective than other optimization algorithms in improving overall performance under varying traffic conditions. Under some isolated conditions, the stability index of merging of the HCOMC may be worse than that of the FIFO model. However, it consistently ensures safe car-following and lane-changing, and improves traffic efficiency, with its stability index only marginally lower than that of the FIFO model. Furthermore, the HCOMC significantly outperforms both the other two on-ramp merging control models in terms of rapidity of merging and fuel consumption economy under most traffic conditions. Considering its comprehensive performance across various metrics under different traffic densities and CAV penetration rates, our HCOMC enhances overall traffic flow performance in the two-lane highway on-ramp merging areas compared to the other two models.

## V. CONCLUSIONS

This paper proposes an innovative HCOMC framework, specifically designed for heterogeneous traffic flow involving CAVs and HDVs on two-lane highways. The HCOMC model consists of three key components: a longitudinal-lateral cooperative planning model, a discretionary lane-changing decision model and a multi-objective optimization model. Finally, comprehensive simulations are conducted to validate the effectiveness of the HCOMC under typical working conditions, i.e. varying traffic densities and CAV penetration rates. The results show that the HCMOC improves the traffic flow efficiency, ensures good fuel consumption economy, and improves the rapidity and stability of merging while maintaining safety standards compared with the benchmarks. These findings highlight the substantial impact of CAV technologies on advanced mixed traffic flow and lay foundation for developing cooperative control strategies tailored to heterogeneous traffic flow to ease traffic congestion and reduce accident risk in bottlenecks.